\documentclass[11pt,a4paper]{article}
\usepackage[hyperref]{acl2021}
\usepackage{times}
\usepackage{latexsym}

\usepackage{microtype}
\usepackage{graphicx}

\aclfinalcopy

\newcommand\blfootnote[1]{%
  \begingroup
  \renewcommand\thefootnote{}\footnote{#1}%
  \addtocounter{footnote}{-1}%
  \endgroup
}

\title{Disentangling Online Chats with DAG-Structured LSTMs}
\author{Duccio Pappadopulo$^{*1}$, Lisa Bauer$^{*2}$, Marco Farina$^1$, Ozan İrsoy$^1$, Mohit Bansal$^2$ \\
  $^1$Bloomberg\quad $^2$UNC Chapel Hill \\
  \texttt{\{dpappadopulo, mfarina19, oirsoy\}@bloomberg.net}\\
  \texttt{\{lbauer6, mbansal\}@cs.unc.edu}}

\date{}

\begin{document}
\maketitle
\begin{abstract}
Many modern messaging systems allow fast and synchronous textual communication among many users. The resulting sequence of messages  hides a more complicated structure in which independent sub-conversations are interwoven with one another. This poses a challenge for any task aiming to understand the content of the chat logs or gather information from them. The ability to disentangle these conversations is then tantamount to the success of many downstream tasks such as summarization and question answering. 
Structured information accompanying the text such as user turn, user mentions, timestamps, is used as a cue by the participants themselves who need to follow the conversation and has been shown to be important for disentanglement.
DAG-LSTMs, a generalization of Tree-LSTMs that can handle directed acyclic dependencies, are a natural way to incorporate such information and its non-sequential nature. In this paper, we apply DAG-LSTMs to the conversation disentanglement task. We perform our experiments on the Ubuntu IRC dataset. We show that the novel model we propose achieves state of the art status on the task of recovering \emph{reply-to} relations and it is competitive on other disentanglement metrics.
\end{abstract}

\section{Introduction}
\label{sec:intro}
\blfootnote{$*$ Equal contribution }
Online chat and text messaging systems like Facebook Messenger, Slack, WeChat, WhatsApp, are common tools used by people to communicate in groups and in real time. In these venues multiple independent conversations often occur simultaneously with their individual utterances interspersed. 

\begin{figure}[t]
    \includegraphics[width=\columnwidth]{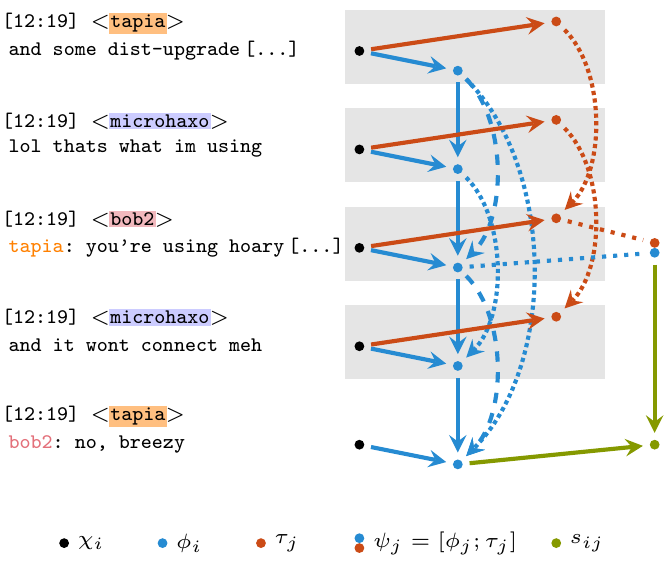}

  \caption{\label{fig:daglstm}    Excerpt from the IRC dataset (\emph{left}) and our reply-to classifier architecture (\emph{right}). 
  Blue dots represent a unidirectional DAG-LSTM unit processing the states coming from the children of the current node. Red dots represent the GRU units performing thread encoding.
  At this point in time, we are computing the score (log-odds) of fifth utterance replying to the third. 
  }
\end{figure}
It is reasonable to assume the existence of an underlying \emph{thread} structure partitioning the full conversation into disjoint sets of utterances, which ideally represent independent sub-conversations. The task of identifying these sub-units, \emph{disentanglement}, is a prerequisite for further downstream tasks among which question answering, summarization, and topic modeling~\cite{traum-etal-2004-evaluation, 10.1145/1148170.1148180, 4597251, elsner2010disentangling}. Additional structure can generally be found in these logs, as a particular utterance could be a response or a continuation of a previous one. Such \emph{reply-to} relationships implicitly define threads as the connected components of the resulting graph topology, and can then be used for disentanglement~\cite{mehri2017chat, dulceanu, wang, guo2018who}.

Modeling work on conversation disentanglement spans more than a decade. \newcite{elsner2008you, elsner2010disentangling} use feature based linear models to find pairs of utterances belonging to the same thread and heuristic global algorithms to assign posts to threads. \newcite{mehri2017chat} and \newcite{jiang2018learning}, while also adopting similar heuristics, use features extracted through neural models, LSTMSs~\cite{hochreiter1997long} and siamese CNNs~\cite{bromley1993signature} respectively. \newcite{10.1145/2009916.2009976} follow a different approach by modeling the interactions between the predicted reply-to relations as a conditional random field.

One challenge in building automatic systems that perform disentanglement is the scarcity of large annotated datasets to be used to train expressive models.
A remarkable effort in this direction is the work of~\newcite{kummerfeld2019large} and the release of a dataset containing more that $77k$ utterances from the IRC \texttt{\#Ubuntu} channel with annotated reply-to structure. In the same paper, it is shown how a set of simple handcrafted features, pooling of utterances GloVe embeddings~\cite{pennington2014glove}, and a feed-forward classifier can achieve good performances on the disentanglement task. Most of the follow-up work on the dataset relies on BERT~\cite{devlin-etal-2019-bert} embeddings to generate utterance representations~\cite{zhu2020did,gu2020pretrained,ti2020dialbert}. \newcite{zhu2020did} use an additional transformer module to contextualize these representations, while \newcite{gu2020pretrained,ti2020dialbert} use an LSTM. 
Two exceptions are~\newcite{ijcai2020-0535}, which models thread membership in an online fashion and discards reply-to relationships, and the recent \newcite{yu2020online} which uses pointer networks~\cite{vinyals2015pointer}.

In this short paper, we use DAG-structured LSTMs~\cite{rsoy2019dialogue} to study disentanglement. As a generalization of Tree-LSTMs~\cite{tai2015improved}, DAG-LSTMs allow to faithfully represent the structure of a conversation, which is more properly described as a directed acyclic graph (DAG) than a sequence. Furthermore, DAG-LSTMs allow for the systematic inclusion of structured information like user turn and mentions in the learned representation of the conversation context. We enrich the representation learned by the DAG-LSTM by concatenating to it a representation of the thread to which the utterance belongs. This \emph{thread encoding} is obtained by means of a GRU unit~\cite{choetal2014properties} and captures thread specific features like style, topic, or persona. Finally we manually construct new features to improve username matching, which is crucial for detecting user mentions, one of the most important features for disentanglement. 

Our results are summarized in Table~\ref{table:results}. The DAG-LSTM significantly outperforms the BiLSTM baseline. Ablation studies show the importance of the new features we introduce. When augmented by thread encoding and a careful handling of posts predicted to be thread starters, the DAG-LSTM architecture achieves state of the art performances on \emph{reply-to} relation extraction on the IRC Ubuntu dataset and it is competitive on the other metrics which are relevant to disentanglement.

\section{Methodology}
\label{sec:methods}

\subsection{Problem Statement}
\label{sec:problemstatement}
A multi-party chat $C$ is a sequence of posts $(c_i)_i$, $i=1, \ldots, |C|$. For each \emph{query} post $c_i$ we look for the set of \emph{link} posts $\mathcal R(c_i)$ such that $c_i$ replies to, or \emph{links to}, $c_j$ for $c_j\in\mathcal R(c_i)$.
When a post $c$ is a conversation starter we define, consistently with~\newcite{kummerfeld2019large}, $\mathcal R(c)=\{c\}$, that is $c$ replies to itself, it is a \emph{self-link}.
This reply-to binary relation defines a DAG over $C$. By taking the union of the reply-to relation with its converse and by calculating its transitive closure, we obtain an equivalence relation on $C$ whose equivalence classes are threads, thus solving the disentanglement problem.

We frame the problem as a sequence classification task. For each query post $c_i$ we consider its $L$ preceding posts $\mathcal O_{c_i}\equiv\{c_{i-L-1},\ldots, c_{i}\}$ and predict one of them as its link. In the IRC Ubuntu dataset, predicting a single link per query post is a good approximation, holding true for more than 95\% of the annotated utterances. We use $L=50$ in the following.
As described in Sections~\ref{sec:context} and \ref{sec:threadenc}, for each query utterance $c_i$, we construct a contextualized representation, $\phi_i \equiv \phi(c_i, C)$. We do the same for each of the links $c_j \in \mathcal O_{c_i}$, using a representation $\psi$ that can in principle differ from $\phi$. We then calculate $p(c_i\,{\textrm{replies-to}}\,c_j)\equiv p(c_j|c_i)$ as
\begin{equation}
    \label{probability}
    p(c_j|c_i)\equiv  \frac{\exp (s_{ij})}{\sum_{c_k \in \mathcal O_{c_i}} \exp (s_{ik})},
\end{equation}
where $s_{ij}\equiv s(\phi_i, \psi_j, f_{ij})$ is a real-valued scoring function described in Section~\ref{sec:scoring} and $f_{ij}$ are additional features. The parameters of the resulting model are learned by maximizing the likelihood associated to Eq.~\ref{probability}. At inference time we predict $\hat j = \textrm{argmax}_{c_j \in \mathcal O_{c_i}} p(c_j|c_i)$.

\subsection{Contextual Post Representation}
\label{sec:context}
The construction of the $\phi$ and $\psi$ representations closely follows~\newcite{rsoy2019dialogue}. Every post $c_i$ is represented as a sequence of tokens $(t_n^i)_n$. An embedding layer maps the tokens to a sequence of $d_I$-dimensional real vectors $(\omega_{n}^i)_n$. We use the tokenizer and the word embeddings from~\newcite{kummerfeld2019large}, $d_I=50$. We generate a representation $\chi_i$ of $c_i$ by means of a single BiLSTM layer unrolled over the sequence of the token embeddings $(\upsilon^i_n)_n\equiv \textrm{BiLSTM}[(\omega_{n}^i)_n]$ followed by elementwise max-affine pooling $\chi_i = \max_{n} {\textrm{Affine}}[(\upsilon^i_n)_n]$.

To obtain the contextualized representations $\phi$, we use a DAG-LSTM layer. This is an N-ary Tree-LSTM~\cite{tai2015improved} in which the sum over children in the recursive definition of the memory cell is replaced with an elementwise max operation~(see Appendix). This allows the existence of multiple paths between two nodes (as it is the case if a node has multiple children) without the associated state explosion~\cite{rsoy2019dialogue}. This is crucial to handle long sequences, as in our case.

At each time step the DAG-LSTM unit receives the utterance representation $\chi_i$ of the current post $c_i$ as the input and all the hidden and cell states coming from a labeled set of children, $\mathcal C(c_i)$, see Figure~\ref{fig:daglstm}. In our case $\mathcal C(c_i)$ contains three elements: the previous post in the conversation ($c_{i-1}$), the previous post by the same user of $c_i$, the previous post by the user mentioned in $c_i$ if any. More dependencies can be easily added making this architecture well suited to handle structured information.
The DAG-LSTM is unrolled over the sequence $(\{\chi_i, \mathcal C(c_i)\})_i$, providing a sequence of contextualized post representations $(\phi_i)_i$. 
We also consider a bidirectional DAG-LSTM defined by a second unit processing the reversed sequence $\tilde c_i\equiv c_{|C|-i+1}$. Forward and backward DAG-LSTM representations are then concatenated to obtain $\phi$.

\subsection{Thread Encoding}
\label{sec:threadenc}
The link post representation $\psi$ can coincide with the query one, $\psi_j\equiv \phi_j$. One potential issue with this approach is that $\psi$ does not depend on past thread assignments. Furthermore, thread-specific features such as topic and persona, cannot be easily captured by the hierarchical but sequential model described in the previous section.
Thus we augment the link representations by means of \emph{thread encoding}~\cite{ijcai2020-0535}. Given a query, $c_i$, and a link $c_j$ posts pair, we consider the thread $\mathcal T(c_j) =(c_{t_i})$, $t_i < t_{i+1}$,  $t_{|\mathcal T(c_j)|}=j$, to which $c_j$ has been assigned. We construct a representation $\tau_j$ of such thread by means of a GRU cell, $\tau_j = {\textrm{GRU}}[(\chi(c))_{c\in \mathcal T(c_j)}]$. $\psi_j$ is then obtained by concatenating $\phi_j$ and $\tau_j$. At training time we use the gold threads to generate the $\tau$ representations, while at evaluation time we use the predicted ones.

\subsection{Scoring Function}
\label{sec:scoring}
Once query and link representations are constructed we use the scoring function in Eq.~\ref{probability} to score each link against the query utterance, with $s$ a three-layer feed-forward neural network. The input of the network is the concatenation $[\phi_i; \psi_j; f_{ij}]$, where $f_{ij}$ are the 77 features introduced by~\newcite{kummerfeld2019large}. We augment them by 42 additional features based on Levenshtein distance and longest common prefix between query's username and words in the link utterance (and viceversa). These are introduced to improve mention detection by being more lenient on spelling mistakes (see~\ref{sec:userfeats} for precise definitions).

\subsection{User Features}
\label{sec:userfeats}
While IRC chats allow systematically tagging other participants (a single mention per post), users can address each other explicitly by typing usernames. This allows for abbreviations and typos to be introduced, which are not efficiently captured by the set of features used by~\newcite{kummerfeld-etal-2019-large}. To ameliorate this problem we construct additional features. Given a pair of utterances $c_1$ and $c_2$ we define the following:
\begin{itemize}
    \item Smallest Levenshtein distance ($D_L$) between $c_1(c_2)$'s username and each of the word in $c_2(c_1)$; 5 bins, $D_L = i$ for $i = 0,\ldots,4$ or $D_L > 4$ .
    \item Largest length of common prefix ($\ell$) between $c_1(c_2)$'s username and each of the word in $c_2(c_1)$; 5 bins, $\ell = i$ for $i = 3,\ldots,6$ or $\ell > 6$.
    \item Binary variable indicating whether $c_1(c_2)$'s username is a prefix of any of the words in $c_2(c_1)$.
\end{itemize}
These amount to a total of 42 additional features for each pair of posts.

\section{Results}

\renewcommand{\arraystretch}{1.2}
\begin{table}
\small
\centering
\begin{tabular}{l @{\hspace{1.2\tabcolsep}} c @{\hspace{0.5\tabcolsep}} c @{\hspace{0.5\tabcolsep}} c @{\hspace{1.2\tabcolsep}} c @{\hspace{0.5\tabcolsep}} c @{\hspace{0.5\tabcolsep}} c}
\hline
\textbf{Model}& \multicolumn{3}{c}{\textbf{Graph}} & \multicolumn{3}{c}{\textbf{Cluster}}\\
 & ${\rm P}$ & ${\rm R}$ & ${\rm F}$ & ${\rm P}$ & ${\rm R}$ & ${\rm F}$\\
\hline
\hline
\citeauthor{kummerfeld2019large}&  $73.7 $ & $ 71.0 $ & $ 72.3$ & $ 34.6$ & $ 38.0$ & $ 36.2$  \\
\citeauthor{zhu2020did}$^\star$ & $73.2$ & $ 69.2$ & $ 70.6$ & $35.8$ & $ 32.7$ & $ 34.2$\\
\citeauthor{ti2020dialbert}$^\star$ &  & & & $42.3$ & $\underline{46.2}$ & $44.1$ \\
\citeauthor{yu2020online} &$74.7$ & $72.7$ & $73.7$& $33.0$ & $38.9$ & $36.0$ \\
 ~$+$~ self.& $\underline{74.8}$ & $\underline{72.7}$ & $\underline{73.7}$ & $42.2$ & $40.9$ & $41.5$\\
~$+$~ joint train, self.& $74.5$ & $71.7$ & $73.1$ & $\underline{44.9}$ & $44.2$ & $\underline{44.5}$ \\
\hline
BiLSTM\,($\downarrow$) &  $73.9$ & $71.2$ & $72.5$ & $31.3$ & $37.5$ & $34.1$ \\
DAG-LSTM & $74.9$ & $72.2$ & $73.6$  & $37.3$ & $42.3$ & $39.6$ \\
~$-$~user features\,($\downarrow$)  & $74.0$ & $71.3$ & $72.6$  & $33.6$ & $39.7$ & $36.4$ \\
~$-$~mention link  & $74.5$ & $71.8$ & $73.1$  & $33.5$ & $38.3$ & $35.7$ \\
~$+$~self.\,($\uparrow$) & $74.9$ & $72.6$ & $73.8$ & $41.1$ & $41.1$ & $41.1$ \\    
~$+$~thread enc. & $75.0$ & $72.3$ & $73.7$ & $37.3$ & $\mathbf{42.5}$ & $39.7$ \\
~$+$~thread enc., self. & $\mathbf{75.2}$ & $ \mathbf{72.7}$ & $ \mathbf{73.9}$ & $\mathbf{42.4}$ & $ 41.7$ & $ \mathbf{42.0}$  \\
\hline
\end{tabular}
\caption{\label{table:results}
Results of our experiments (\emph{bottom}, best in bold) and literature (\emph{top}, best underlined). The $\uparrow$($\downarrow$) sign indicates the model being significantly better (worse) ($\textrm{p}<0.05$) than the DAG-LSTM entry based on a McNemar test~\cite{mcnemar1947note} conducted on the \emph{test} set. User features and mention links are included in this baseline model, thread encoding and self-link threshold tuning are not. Starred entries use contextual embeddings.
}
\end{table}

\subsection{Evaluation}
\label{sec:evaluation}
We conduct our experiments on the Ubuntu IRC dataset for disentanglement \cite{kummerfeld2019large, kim2019eighth}. We focus on two evaluation metrics defined in~\newcite{kummerfeld2019large}: \emph{graph} F$_1$, the F-score calculated using the number of correctly predicted reply-to pairs; \emph{cluster} F$_1$, the F-score calculated using the number of matching threads of length greater than 1.

\subsection{Experiments}
\label{sec:architectures}
As a baseline, we use a BiLSTM model in which $\phi_i(=\psi_i)$ is obtained as the hidden states of a bidirectional LSTM unrolled over the sequence $(\chi_i)_i$.
The base DAG-LSTM model uses both username and mentions to define the children set $\mathcal C$ of an utterance. Bidirectionality is left as a hyperparameter.
All our experiments use the same architecture from section~\ref{sec:methods} to construct the utterance representation $\chi$.
We train each model by minimizing the negative log-likelihood for Eq.~\ref{probability} using Adam optimizer~\cite{kingma2019method}. We tune the hyperparameters of each architecture through random search.\footnote{We refer to the Appendix for details.}
Table~\ref{table:results} shows the test set performances of the models which achieve the best graph ${\textrm{F}}_1$ score over the dev set. Optimizing graph over cluster score is motivated by an observation: dev set cluster ${\textrm{F}}_1$ score displays a much larger variance than graph ${\textrm{F}}_1$ score, which is roughly four-fold after subtracting the score rolling average. By picking the iteration with the best cluster ${\textrm{F}}_1$ score we would be more exposed to fluctuation and to worse generalization, which we observe.

\subsection{Self-Links Threshold Tuning}
As noted by \newcite{yu-joty-2020-online}, the ability of the model to detect self-links is crucial for its final performances. In line with their findings, we also report that all our models are skewed towards high recall for self-link detection (Table~\ref{table:selflinks}).

\renewcommand{\arraystretch}{1.2}
\begin{table}
\small
\centering
\begin{tabular}{l @{\hspace{1.2\tabcolsep}} c @{\hspace{0.5\tabcolsep}} c @{\hspace{0.5\tabcolsep}} c }
\hline
\textbf{Model}& \multicolumn{3}{c}{\textbf{Self-links}} \\
 & ${\rm P}$ & ${\rm R}$ & ${\rm F}$ \\
\hline
\hline
BiLSTM & $79.6$ & $94.6$ &  $86.5$ \\
DAG-LSTM & $82.8$&$93.8 $&$88.0$ \\
~$+$~self-links threshold &  $87.7$ &$92.4$& $90.0$ \\    
DAG-LSTM + thread enc. & $81.4$ & $93.8$ & $87.2$  \\
~$+$~self-links threshold &$89.8$& $90.6$ & $90.2$ \\    
\hline
\end{tabular}
\caption{\label{table:selflinks}
Thread starters (\emph{self-links}) performances for our models in Table~\ref{table:results}, before and after thresholding.
}
\end{table}

To help with this, we introduce two thresholds $\theta$ and $\delta$, which we compare with $\hat p$, the argmax probability Eq.~\ref{probability}, and $\Delta p$, the difference between the top-2 predicted probabilities. Whenever the argmax is a self-link: if $p<\theta$, we predict the next-to-argmax link, otherwise we predict both the top-2 links if also $\Delta \hat p < \delta$. On the dev set, we first fine-tune $\theta$ to maximize the self-link F$_1$ score and the fine-tune $\delta$ to maximize the cluster F$_1$ score.

\subsection{Results Discussion}
\label{sec:results}
Table~\ref{table:results} shows our main results.  Our DAG-LSTM model significantly outperforms the BiLSTM baseline. We perform ablation studies on our best DAG-LSTM model showing that while both user features and mention link provide a performance improvement for both cluster and graph score, only user features ablation results in a significant change. Self-links threshold tuning improves performances, particularly on cluster score for both models, highlighting the importance of correctly identifying thread starters. 

The DAG-LSTM model with thread encoding achieves state of the art performances in predicting $\emph{reply-to}$ relations. This is particularly interesting especially when we compare with models employing contextual embeddings like \newcite{zhu2020did}. For the cluster scores, the best model is the pointer network model of \newcite{yu2020online}, which is anyway within less than 0.5\% of the best contextual model, and within 2.5\% of our model.
The difference mainly arises from a difference in recall and corresponds to an absolute difference of less than 10 true positive clusters on the test set. Further comparisons with existing literature are limited by code not being available at the moment.

\section{Conclusions}
\label{sec:conclusions}

In this paper we apply, for the first time, DAG-LSTMs to the disentanglement task; they provide a flexible architecture that allows to incorporate into the learned neural representations the structured information which comes alongside multi-turn dialogue. We propose thread encoding and a new set of features to aid identification of user mentions.

There are possible directions left to explore. We modeled the reply-to relationships in a conversation by making an assumption of conditional independence of reply-to assignments. This is possibly a poor approximation and it would be interesting to lift it. A challenge with this approach is the computational complexity resulting from the large dimension of the output space of the reply-to classifier. We notice that thread encoding allows a non-greedy decoding strategy through beam search which would be interesting to further explore.

\section*{Acknowledgments}
We thank the reviewers for their useful feedback. We thank Andy Liu, Camilo Ortiz, Huayan Zhong, Philipp Meerkamp, Rakesh Gosangi, Raymond (Haimin) Zhang for their initial collaboration. We thank Jonathan Kummerfeld for discussions. This work was performed while Lisa interned at Bloomberg, and was later supported by DARPA MCS Grant N66001-19-2-4031, NSF-CAREER Award 1846185, and a NSF PhD Fellowship. The views are those of the authors and not of the funding agencies.

\bibliography{anthology,acl2021}
\bibliographystyle{acl_natbib}
ssss
\appendix
\section{Appendix}

\subsection{DAG-LSTM Equations}
\label{sec:daglstm}
A DAG-LSTM is a variation on the Tree-LSTM~\cite{tai-etal-2015-improved}
architecture, that is defined over DAGs. Given a DAG, $G$, we assume that for every vertex $v$ of $G$, the edges $e(v, v')$ connecting the children $v' \in \mathcal C(v)$ to $v$ can be assigned a unique label $\ell_{v, v'}$ from a fixed set of labels.

A pair of states vectors $(h_v, c_v)$ and an input $x_v$ are associated to every vertex $v$. The DAG-LSTM equations define the states $(h_v, c_v)$, as a function of the input $x_v$ and the states of its children:
\begin{equation}
    (h_v, c_v) = {\textrm{DAG-LSTM}}(x_v; \{(h_w, c_w)|w \in \mathcal C(v)\}).
\end{equation}
The equations defining such functions are the following:
\begin{eqnarray}
i_v &=& \sigma\bigg(W_{ix}x_v+\sum_{v'\in \mathcal C(v)}W^{\ell_{v, v'}}_{ih} h_{v'}\bigg) \\ 
f_{vv'} &=& \sigma\bigg(W_{fx}x_v+\sum_{v''\in \mathcal C(v)}W^{\ell_{v, v'}\ell_{v, v''}}_{fh} h_{v''}\bigg) \phantom{xxx}\\ 
c_v &=& i_v\odot u_v + \max_{v'\in \mathcal C(v)} f_{vv'}\odot c_{v'}\label{dagmax}\\
h_v &=& o_v \odot \tanh(c_v)
\end{eqnarray}
The equations for the $o$ and $u$ gates are the same as those for the $i$ gate by replacing everywhere $i\to o, u$.
Bias vectors are left implicit in the definition of $i$, $f$, $o$, and $u$. $\odot$ represents Hadamard product and ${\textrm{max}}$ in Eq.~\ref{dagmax} represent elementwise max operation.

A bidirectional DAG-LSTM, is just a pair of independent DAG-LSTM, one of which is unrolled over the time reversed sequence of utterances. The output of a bidirectional DAG-LSTM is the concatenation of the $h$ states of the forward and backward unit for a given utterance.

\subsection{Training and Hyperparameter Tuning}
We use adjudicated training, development, and test sets from~\cite{kummerfeld-etal-2019-large}. Each of these dataset is composed a set of conversation (153 in the training set and 10 in both development and test set) each representing a chunk of contiguous posts from the IRC \texttt{\#Ubuntu} channel. Each of these conversation contains strictly more than 1000 posts (exactly 1250 and 1500 for dev and test set respectively). Annotations are available for all but the first 1000 posts in every conversation. We apply some preprocessing to these conversations. We chunk the annotated section of every training conversation in contiguous chunks of 50 posts each, starting from the first annotated post.~\footnote{This may result in the last chunk to have less than 50 posts. This happens for 45 conversations.} To each of these chunks we attach a past context of 100 posts and a future context of 50, resulting in 200 utterances long chunks. For each of these chunks we keep only those annotated links for which the response utterance lies in the central 50 posts. We do not chunk development and test set, but drop the first 900 post in every conversation.

The various architectures we consider share the same set of parameters to fine-tune. One parameter $d_h$ controls the dimension of the hidden state of the LSTMs and one parameter $d_{FF}$ controls the dimension of the hidden layers of the feed-forward scorer.
We use word dropout, apply dropout after the max-affine layer, and apply dropout after activation at every layer of the feed-forward scorer. We clip all gradient entries at 5. We use a single layer of LSTMs and DAG-LSTMs to build the $\chi$ and $\phi,\psi$ representations and we do not dropout any of their units. Similarly we use a single layer GRU for the thread encoder.
We list all the hyperparameters in Table~\ref{table:ht} together with their range and distribution used for the random search.

Hyperparameter optimization is performed by running 100 training jobs for the base BiLSTM architecture, DAG-LSTM, and DAG-LSTM with thread encoding. Our published results are from the best among these runs. The best sets of parameters we find for each of these architectures are:
\begin{itemize}
    \item BiLSTM: $d_h=256$, $d_{FF}=128$, no word and max-affine dropout, a feed forward-dropout equal to 0.3, and a learning rate of $2.4\times 10^{-4}$.
    \item DAG-LSTM: $d_h=64$, $d_{FF}=256$, no word and max-affine dropout, a feed forward-dropout equal to 0.3, and a learning rate of $7.3\times 10^{-4}$.
    \item DAG-LSTM with thread encoding: $d_h=d_{FF}=256$, word and max-affine dropout equal to 0.3, a feed forward-dropout equal to 0.5, and a learning rate of $7.9\times 10^{-4}$.
\end{itemize}

User feature and mention link ablations are obtained by fixing all parameters of the best DAG-LSTM run (removing the feature we are experimenting with) and running 10 jobs by only changing the random seed.

Each training job is performed on a single GPU and, depending on the architectures, takes from 6 to 12 hours.

\renewcommand{\arraystretch}{1.2}
\begin{table}
\small
\centering
\begin{tabular}{lcc}
\hline
Parameter & Domain & Distribution\\
\hline
\hline
$d_{h}$& $\{64, 128, 256\}$ & categorical \\ 
$d_{FF}$& $\{64, 128, 256\}$ & categorical \\ 
word dropout& $\{0, 0.3, 0.5\}$ & categorical \\
max-affine dropout& $\{0, 0.3, 0.5\}$ & categorical \\
feed-forward dropout& $\{0, 0.3, 0.5\}$ & categorical \\
learning rate& $[10^{-5}, 10^{-3}]$ & log-uniform \\
BiDAG-LSTM& $\{{\textrm{true}, \textrm{false}}\}$ & categorical \\
\hline
\end{tabular}
\caption{\label{table:ht} Hyperparameters of the model architectures. During hyperparameter optimization, we perform a random search according to the distributions described above. Categorical distributions have uniform probability mass function.}
\end{table}

\subsection{Significance Estimates}
\label{sec:significance}
We use McNemar test~\cite{mcnemar1947note} to evaluate the significance of performance differences between model. 
Given two models $M_A$ and $M_B$, we define $n_{A\overline B}$ as the number of links correctly predicted by $A$ but not by $B$. Under the null hypothesis both $n_{A\overline B}\sim {\textrm{Bin}}(n_{A\overline B}, n, 1/2)$,  where $n\equiv n_{A\overline B} + n_{B\overline A}$. We define a model $A$ to be \emph{significantly} better than a model $B$ if the null hypothesis is excluded at 95\% confidence level.

\end{document}